\documentclass[a4paper,twoside]{article}

\usepackage{epsfig}
\usepackage{adjustbox}
\usepackage{booktabs}
\usepackage{subcaption}
\usepackage{calc}
\usepackage{amssymb}
\usepackage{amstext}
\usepackage{amsmath}
\usepackage{amsthm}
\usepackage{multicol}
\usepackage{pslatex}
\usepackage{apalike}
\usepackage{xcolor}
\usepackage{pifont}
\usepackage{lipsum}
\usepackage{multicol}
\usepackage{graphicx}
\usepackage{wrapfig}
\usepackage{SCITEPRESS}
\usepackage[a4paper,
            left=1.025in,
            right=1.025in,
            top=1.3in,
            bottom=1.655in]{geometry}
\usepackage[pagebackref=false,breaklinks=true,letterpaper=true,colorlinks=true,citecolor=black,bookmarks=false]{hyperref}

\definecolor{darkgreen}{RGB}{0, 150, 0}
\definecolor{darkred}{RGB}{200, 0, 0}
\newcommand{\ch}{{\color{darkgreen} \ding{51}}}
\newcommand{\xm}{{\color{darkred} \ding{55}}}
\newcommand{\sy}[1]{}

\begin{document}

\title{LiMoSeg: Real-time Bird's Eye View based LiDAR Motion Segmentation}



\author{\authorname{Sambit Mohapatra\sup{1}, Mona Hodaei\sup{1}, Senthil Yogamani\sup{2}, Stefan Milz\sup{3},
Heinrich Gotzig\sup{1},
Martin Simon\sup{1,4},
Hazem Rashed\sup{1},
Patrick Maeder\sup{4}}

\affiliation{\sup{1}Valeo, Germany \hspace{0.5cm} \sup{2}Valeo, Ireland \hspace{0.5cm}
\sup{3}Spleenlab.ai, Germany \hspace{0.5cm}
\sup{4}TU Ilmenau, Germany}
\email{\{sambit.mohapatra, senthil.yogamani, heinrich.gotzig, martin.simon, hazem.rashed\}@valeo.com, m.hodaei96@gmail.com, stefan.milz@spleenlab.ai, patrick.maeder@tu-ilmenau.de}
\vspace{-1cm}
}
\abstract{Moving object detection and segmentation is an essential task in the Autonomous Driving pipeline. Detecting and isolating static and moving components of a vehicle's surroundings are particularly crucial in path planning and localization tasks. This paper proposes a novel real-time architecture for motion segmentation of Light Detection and Ranging (LiDAR) data. We use three successive scans of LiDAR data in 2D Bird's Eye View (BEV) representation to perform pixel-wise classification as static or moving. Furthermore, we propose a novel data augmentation technique to reduce the significant class imbalance between static and moving objects. We achieve this by artificially synthesizing moving objects by cutting and pasting static vehicles. We demonstrate a low latency of 8 ms on a commonly used automotive embedded platform, namely Nvidia Jetson Xavier. To the best of our knowledge, this is the first work directly performing motion segmentation in LiDAR BEV space. We provide quantitative results on the challenging SemanticKITTI dataset, and qualitative results are provided in \url{https://youtu.be/2aJ-cL8b0LI}.}

\onecolumn \maketitle \normalsize \setcounter{footnote}{0} \vfill
\section{{INTRODUCTION}}
\label{sec:introduction}

Autonomous Driving tasks such as perception which involves object detection~\cite{rashed2021generalized}, \cite{dahal2021online}, \cite{rashedfisheyeyolo}, soiling detection~\cite{uricar2021let}, \cite{das2020tiledsoilingnet}, road edge detection \cite{dahal2021roadedgenet}, weather classification \cite{dhananjaya2021weather}, depth prediction \cite{kumar2021fisheyedistancenet++}, \cite{kumar2018near} are challenging due to the highly dynamic and interactive nature of surrounding objects in the automotive scenarios~\cite{kia_2021}. Identification of the environmental objects as moving and static is crucial to achieving safe motion planning and navigation. An autonomous vehicle's route has to consider future coordinates and velocities of surrounding moving objects. In addition, this information is a critical source for simultaneous localization and mapping (SLAM) \cite{gallagher2021hybrid} and pose estimation \cite{kumar2020fisheyedistancenet}. As the vehicle is in motion, it is difficult to distinguish between background and other moving objects. Thus, motion segmentation requires estimation of the vehicle's ego-motion and compensation to extract other moving objects in the scene. Motion cues can also be used to detect generic moving objects like animals which are difficult to train based on appearance due to their rare appearance in driving scenes and due to their diversity. Relative to appearance-based object detection and semantic segmentation, CNN-based motion segmentation approaches are relatively less mature~\cite{kumar2021omnidet}, ~\cite{yahiaoui2019fisheyemodnet}. \par

Autonomous vehicles are equipped with a variety of sensors to generate an understanding of environments. The most common ones are cameras and LiDAR. Although cameras provide rich color information, they face a lack of depth information and rely on illumination, making them vulnerable to poor illumination conditions such as nights or rainy days and are also prone to adversarial attacks \cite{sobh2021adversarial}. However, providing accurate 3D depth~\cite{kumar2021svdistnet}, \cite{kumar2020unrectdepthnet} and geometric information \cite{kumar2020syndistnet} of the environment without dependency on weather and illumination is possible with LiDAR \cite{kumar2018monocular}. Considering the benefits of LiDAR data, we focus our efforts towards motion segmentation in LiDAR point clouds, building upon the blocks presented in~\cite{mohapatra2021bevdetnet}. A summary of the contributions of this work are listed below:
\begin{itemize}
  \item We propose a novel method to implement real-time motion segmentation on LiDAR point clouds. First, we convert LiDAR 3D point clouds to 2D Bird’s Eye View (BEV), then we classify each pixel of the BEV as static or motion. We demonstrate real-time inference on an embedded GPU platform.
  \item We introduce a novel residual computation layer that directly leverages the motion across frames to increase the disparity between pixel values for static and moving parts of the BEV frames.
  \item We introduce a data augmentation technique to simulate motion by selectively translating static objects across successive frames. The technique addresses the problem of significant class imbalance present in the dataset. 
\end{itemize}
\begin{figure*}[t]
\captionsetup{singlelinecheck=false, font=small, belowskip=-8pt}
\includegraphics[width=\textwidth, height=0.40\textwidth]{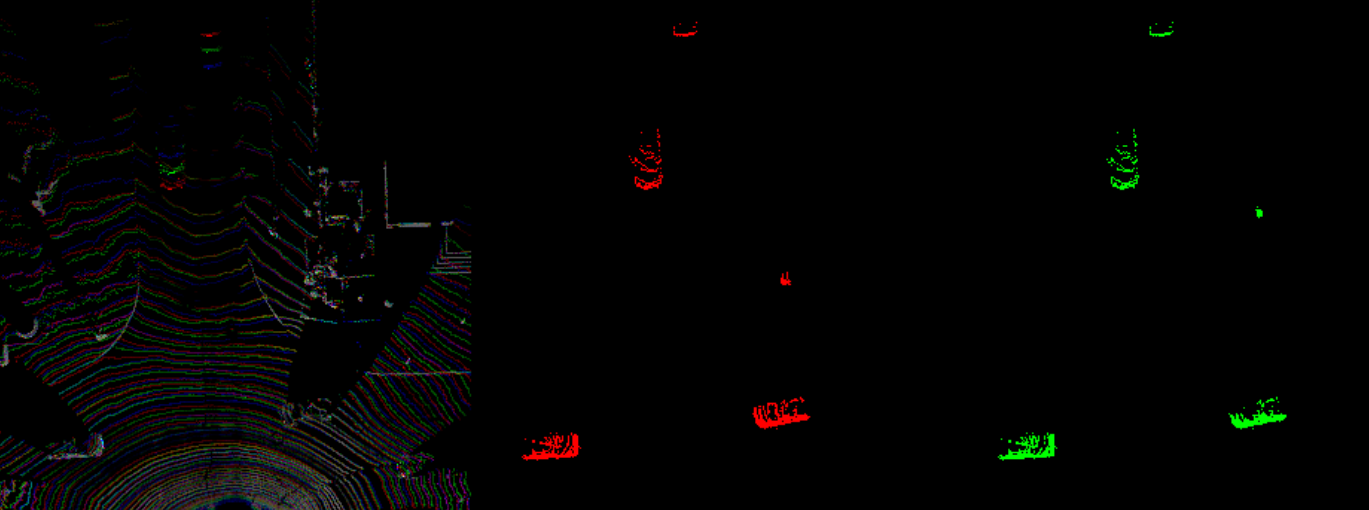}\centering
    \caption{{\textbf{Input 3 frames superimposed - red pixels are present frame (Left), Ground truth motion mask (Middle), Predicted motion mask (Right).}}}
    \centering
    \label{fig:teaser}
\end{figure*}
\section{RELATED WORK}

A variety of approaches have been proposed for moving semantic segmentation using vision~\cite{barnes2018driven}~\cite{patil2020end}~\cite{mcmanus2013distraction}. Apart from vision-based methods, other approaches rely on the fusion of vision and LiDAR sensors~\cite{rashed2019fusemodnet}~\cite{el2019rgb}~\cite{yan2014automatic}~\cite{postica2016robust}. Using LiDAR sensors individually in order to perform semantic segmentation tasks has been taken into consideration recently~\cite{cortinhal2020salsanext}~\cite{li2020multi}~\cite{milioto2019rangenet++}.

Motion segmentation can be performed by LiDAR-based methods based on clustering approaches such as~\cite{dewan2016motion}~including point motion prediction by RANSAC and clustering objects. Vaquero et al.~\cite{vaquero2017deconvolutional} performed motion segmentation after clustering vehicles points and matching objects in consecutive frames. Steinhauser et al.~\cite{steinhauser2008motion} have devised another method to classify moving and non-moving objects, using RANSAC and extracting features from two sequential frames, although in some scenarios like when a vehicle is surrounded by many moving objects or dense trees, the RANSAC algorithm and feature extraction have failed. 

In other studies, Wang et al.~\cite{wang2012could} have segmented objects that are able to move into different categories such as cars, bicycles, and pedestrians in laser scans of urban senses. Consistent temporal information of consecutive LiDAR scans has been utilized with semantic classification and semantic segmentation approaches~\cite{dewan2020deeptemporalseg}~\cite{dewan2017deep} which are developed based on motion vectors of rigid bodies that have been estimated by a flow approach on LiDAR scans~\cite{dewan2016rigid}.~However, distinguishing scene flow from noise in the case of slowly moving objects can be a difficult task to perform. In urban scenarios, most semantic segmentation methods are able to recognize objects typically being in the pedestrians, bicyclists, and cars classes in point clouds~\cite{alonso20203d}~\cite{milioto2019rangenet++}~\cite{wu2019squeezesegv2}~\cite{wu2018squeezeseg}~\cite{biasutti2019lu}~\cite{cortinhal2020salsanext}.~However, none of them distinguish between static and moving objects. There are some studies in order to distinguish moving objects. Yoon et al.~\cite{yoon2019mapless} propose a ray-tracing method including a clustering step to detect moving objects in LiDAR scans, which can, however, occasionally result in incomplete detection of objects or detection of static areas. Shi et al.~\cite{shi2020spsequencenet}~introduce a method based on utilizing sequential point clouds to achieve the prediction of moving objects.
\begin{figure*}[t]
\captionsetup{singlelinecheck=false, font=small, belowskip=-8pt}
\includegraphics[width=\textwidth, height=0.85\textwidth]{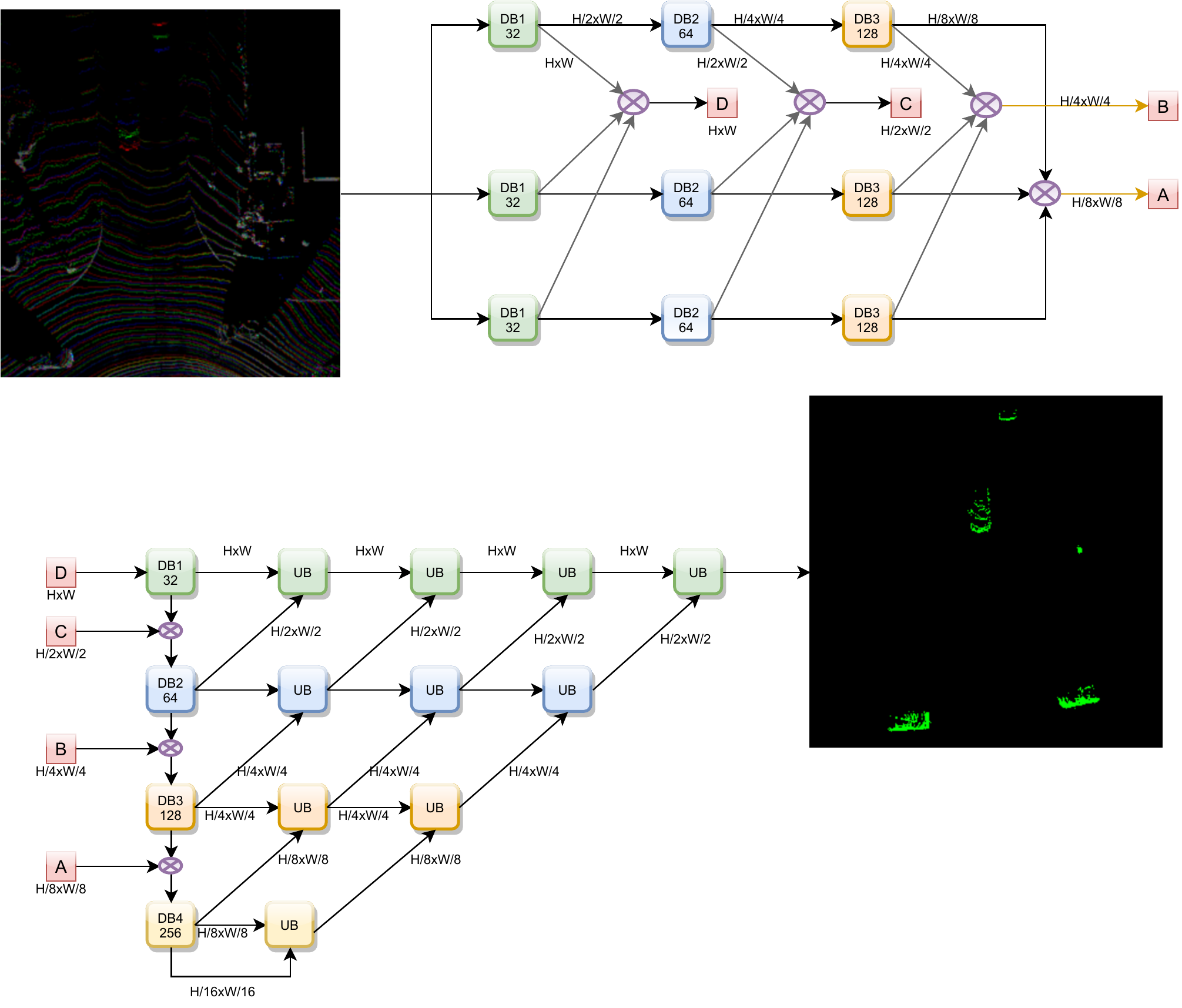}
\centering
    \caption{\textbf{Overall architecture of LiMoSeg} with individual encoders (top) and joint encoder and decoder parts (bottom). The three input BEV frames are superimposed as red-greed-blue channels. The red channel is the present frame, and the green and blue channels are past two frames in sequence.}\centering
    \label{fig:limoseg-arch}
\end{figure*}
However, most of the used architectures were primarily developed for semantic segmentation and have a relatively large number of parameters.
\begin{figure*}[t]
\captionsetup{singlelinecheck=false, font=small, belowskip=-8pt}
\includegraphics[width=\textwidth, height=0.40\textwidth]{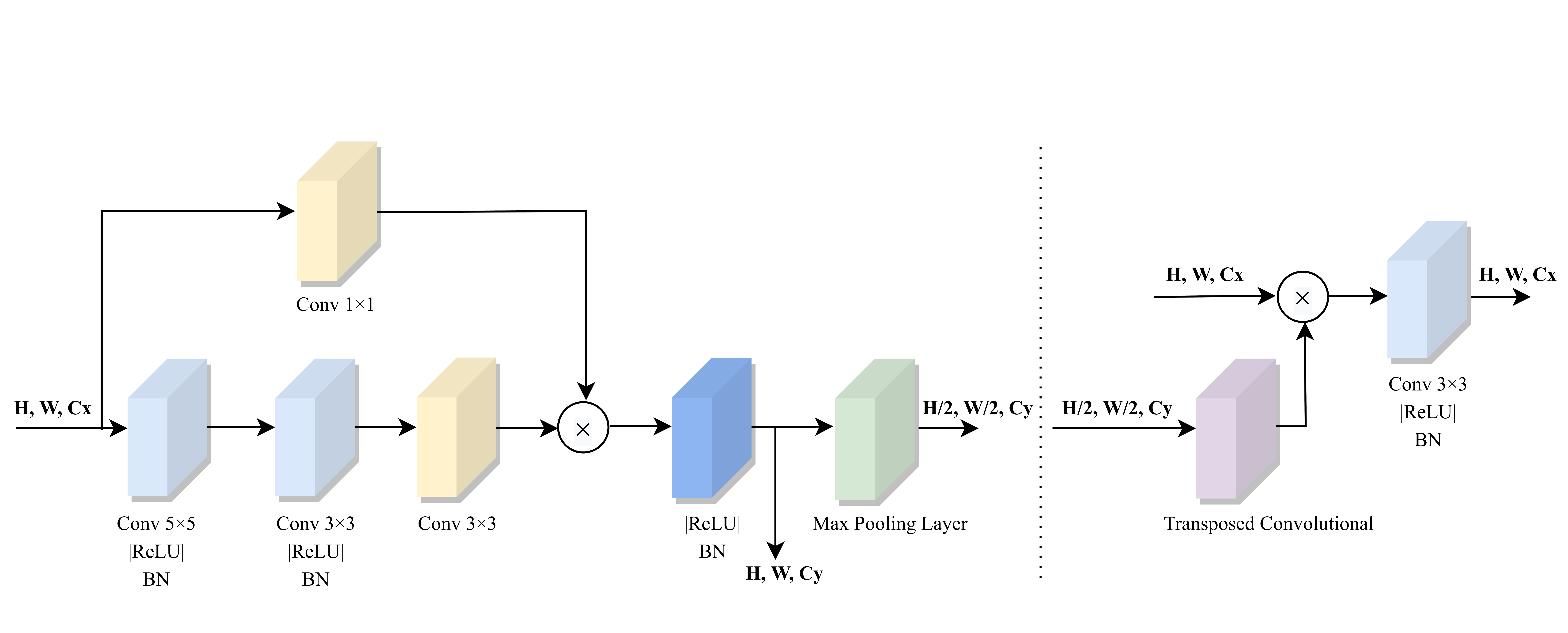}\centering
    \caption{\centering\textbf{Downsampling Block (DB) (Left) and Upsampling Block (UB) (Right).}}
    \label{fig:db}
\end{figure*}
\section{PROPOSED METHOD}
\subsection{Input Data Preparation}

Since a set of sequential point clouds is needed for motion segmentation, we group each frame in the SemanticKITTI~\cite{behley2019iccv} dataset with its past two frames (we use 2 past frames) in the sequence. The past frames are then motion compensated using the pose matrices provided in the dataset as described by the Eq.~\ref{mocomp}. The motivation behind using 2 past frames is to strike a balance between the accuracy of predictions and network size and latency. The number of encoders is directly proportional to the number of input frames.
\begin{equation} 
\label{mocomp}
    F_{N-1}^{N} = P_{N}^{-1} \cdot (P_{N-1} \cdot F_{N-1})
\end{equation}

where
\\
\(F_{N}: frame \: N\\
F_{N-1}: frame \: N-1\\
P_{N}: Pose \: matrix \: of \: frame \: N\\
P_{N-1}: Pose \: matrix \: of \: frame \: N-1\) \\

The set of motion-compensated frames are then converted to 2D BEV using the same process as described in \cite{mohapatra2021bevdetnet}, and \cite{barrera2020birdnet+}.
We limit the x and y ranges to (0, 48) and (-16, 16) meters, respectively, for BEV creation.
Using a cell resolution (in BEV space) of 0.1 meters, we generate (480x320) sized BEV images for each LiDAR frame.\par

Unlike \cite{chen2021moving}, we do not use the range image representation. Our reason is that range images are only better at short range due to their skewed aspect ratio. Neighboring pixels in range representations disregard the metric distance from the underlying points compared to BEV.
Objects far away from the car are barely visible in a range image. Furthermore, range images are affected by even partial occlusion. BEV representation overcomes these problems to some extent (particularly for semi-occluded objects) while presenting the benefits of 2D representation. Another key advantage of BEV representation is that reconstruction of 3D points is a simple matter of looking up the row and column indices of the pixels and multiplying by the cell resolution. Furthermore, most downstream applications such as motion planning are made on a grid-based BEV space, and hence predictions available directly in BEV space reduce the number of interconversions.
\subsection{Data Augmentation}

{Due to the rather difficult and expensive process of collecting and annotating LiDAR data, data augmentation has been a key technique used to increase the training set size and also allow better generalization of the network to different scenarios. One of the most commonly used techniques is sampling-based augmentation or ground truth augmentation as described in the Second algorithm \cite{yan2018second}. The idea is to copy objects from some frames and paste them into others, increasing the number and type of objects.
We use this idea but modify it to fit our case of generating artificially moving objects in frames.}

For each frame (3D LiDAR) with no moving objects (no points marked as moving), we collect all the points belonging to class cars. A uniform random value then translates these points along x and y axes for N successive frames, and the translations increase along the x-axis in each frame to produce a notion of motion. Experimentally, we found N=4 to perform best. The transformed points are then concatenated to the rest of the points, and their labels are marked as moving cars. We do not apply this technique to frames with motion objects to avoid clutter due to overlap between the synthetic objects and the actual objects in the frame.
Though simple, this method enables us to make better use of a sizeable portion of the dataset that has no or very few numbers of LiDAR points marked as moving objects.
\subsection{Network Architecture}

{Our goal is to design a pixel-wise prediction model which operates in BEV space, and it is also efficient with very low latency. BEVDetNet \cite{mohapatra2021bevdetnet} is a recent efficient model which operates on BEV space. It produces outputs in the same spatial resolution as input and has a head that does binary keypoint classification. However, for predicting relative motion between frames and classifying each pixel as moving or static, we adapt this architecture.} We inherit the building blocks from BEVDetNet~\cite{mohapatra2021bevdetnet} and build a multi-encoder joint-decoder architecture as shown in Figure \ref{fig:limoseg-arch}. 

The feature extraction blocks are called Downsampling Blocks (DB), as seen in Figure \ref{fig:db}. They use $5\times5$ and $3\times3$ convolutions to capture features at different scales and also successively reduce the spatial resolution of input using a max-pooling operation at the end. The Upsampling Blocks (UB), as seen in Figure \ref{fig:db} are used to increase the spatial resolution of inputs and serve to produce final output at the same spatial dimension as the input. It consists of a transposed convolution followed by a single convolutional block.
We use the ReLU activation function throughout the network. Since we have 3 input BEV images (present frame and past 2 frames), we have 3 individual encoders consisting of 3 DB blocks. The individual encoders compute per-input features. Features from multiple stages of the network are then collected for all three encoders and then fused using a concatenation and multiplication-based fusion approach. A joint feature computation chain consisting of 4 DB blocks then computes joint features upon the pre-computed individual features from each encoder. The idea is individual encoders compute features for objects while the joint encoder computes features that capture the interaction between objects from all three streams. These are essentially the features that capture the relative displacement between objects due to motion. While concatenation is used as a most common feature fusion approach, we augment it by explicitly multiplying features channel-wise from corresponding stages to compute a set of features that forms a loose correlation between similar features across the three channels. We then concatenate these features to the rest of the concatenated features. We use only 3 DB blocks in the individual encoders since their primary job is to compute the low-level features that characterize objects. The joint encoder has 4 DB blocks to allow computation of sufficiently complex features to capture motion. Our motivation was to limit the number of parameters as much as possible.

\subsubsection{Residual Computation Layer}

\begin{figure}[t]
\includegraphics[width=0.50\textwidth, height=0.50\textwidth]{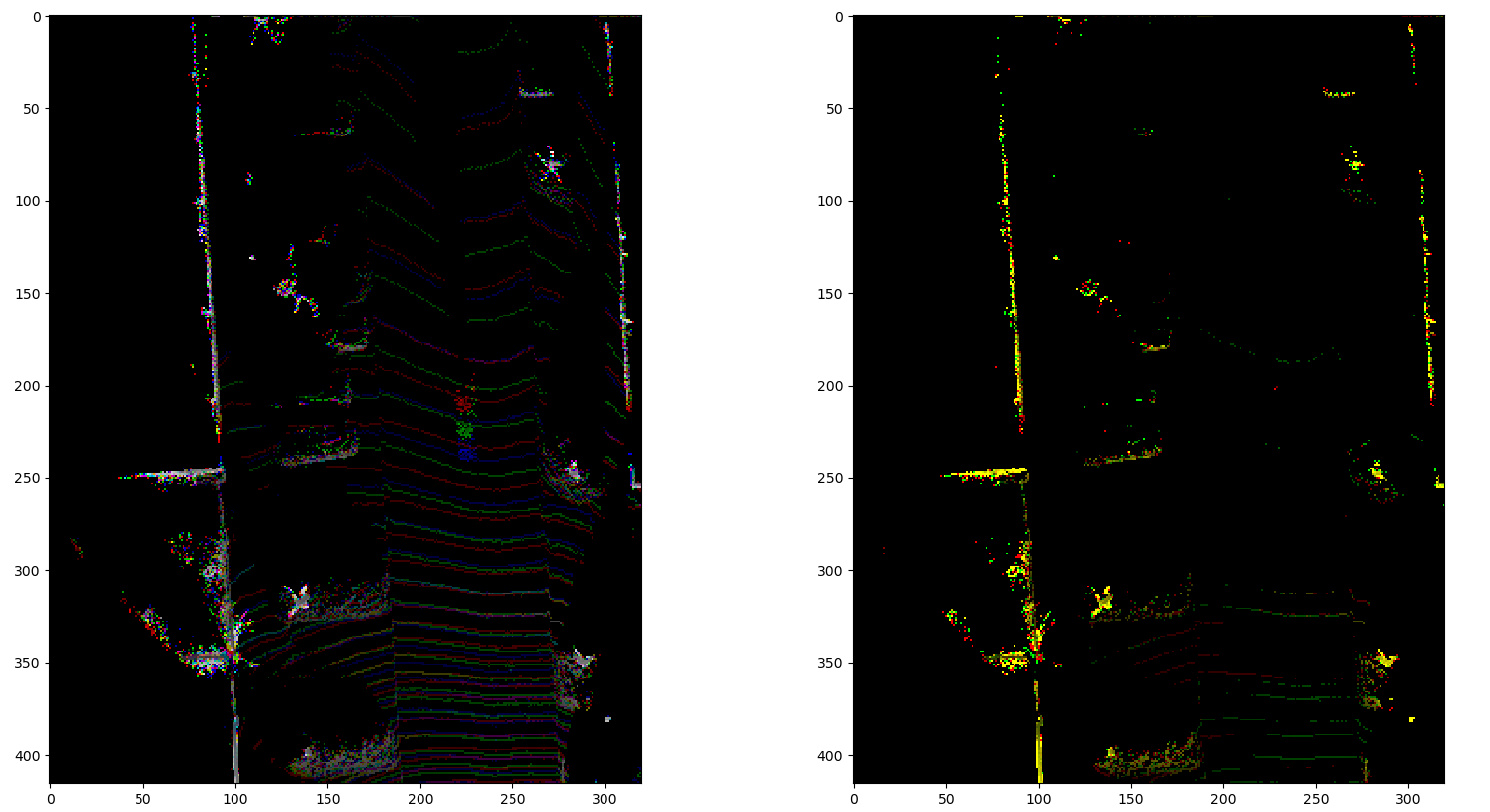}\centering
    \caption{\bf{Three successive frames show a moving object at the center in red-green-blue (Left), Computed residuals with static parts in bright, and mobile objects region in dark at the center (Right).}}
    \centering
    \label{fig:residuals-motion}
\end{figure}

{Residual layers have proven to improve motion segmentation significantly as demonstrated by \cite{chen2021moving}. They compute the difference between successive motion-compensated frames and then normalize it. This produces a disparity map between the moving and static parts of the two frames.
However, we multiply motion-compensated successive frames to generate residuals. Since static objects will have overlaps (some of the parts at least) across successive frames, the residual gets large values in such parts.
Moving objects occupy different spatial locations across successive frames (with some overlap depending on the amount of motion). Therefore, such locations become 0 due to the multiplication as seen in Figure \ref{fig:residuals-motion}. This creates a much more significant disparity in the static and moving parts of the frame and provides a weak attention mechanism to the network. We normalize again after residual computation.}
\vspace{-4pt}
\section{RESULTS}

We follow the standard training and validation split for the SemanticKITTI dataset and keep sequence 8 for validation. All other sequences between [00-10] are converted into BEV, and the motion annotations are also converted into BEV. It was experimentally found that due to the significant class imbalance between frames with motion points and frames without motion points, using all the frames affects the network's performance negatively by acting as a bias towards classifying points as static more often.
\begin{table}[t]
\centering
\caption{\bf Comparison of accuracy and inference latency per frame in BEV for motion segmentation.}
\begin{adjustbox}{width=\columnwidth}
\label{tab:kpibenchmark}
\begin{tabular}{@{}lcc@{}}
\toprule
\textit{\textbf{Method}} & \textit{\begin{tabular}[c]{@{}c@{}}Inference\\ Latency (ms)\end{tabular}} & \textit{\begin{tabular}[c]{@{}c@{}}IoU \\ (moving class)\end{tabular}} \\ 
\midrule
RangeNet++ \cite{chen2021moving} & 45 & 39.5 \\
MINet \cite{chen2021moving}      & 24 & 36.9 \\
SalsaNext \cite{chen2021moving}  & 41 & \textbf{53.4} \\
\textbf{LiMoSeg (Ours)}          & \textbf{8}  & 52.6 \\ 
\bottomrule
\end{tabular}
\end{adjustbox}
\end{table}
Therefore, we only convert those frames to BEV, which have at least 20 motion points in them. We train with a batch size of 12 for only 30 epochs. Since this is a segmentation task, we use weighted cross-entropy loss as our loss function as described by equation \ref{eq_loss}, where $y_c$ is the ground truth class and $\hat{y_c}$ is the predicted class for each pixel in the BEV.
\begin{equation} 
\label{eq_loss}
    L_{motion} = -\sum_{c=1}^{C}w_cy_c\log{\Hat{y_c}}, \quad w_c = \frac{1}{\log(f_c + \epsilon)}
\end{equation}

\begin{table}[t]
\centering
\vspace{0.05in}
\caption{\textbf{Ablation study of different architectural and residual computation settings.}}
\label{tab:ablation1}
\begin{tabular}{@{}lccc@{}}
\toprule
\textit{\textbf{Architecture}} 
& \textit{\begin{tabular}[c]{@{}c@{}}ch. wise \\ mul \end{tabular}} 
& \textit{\begin{tabular}[c]{@{}c@{}}ch. wise \\ sub \end{tabular}} 
& \textbf{IoU} \\ 
\midrule
\begin{tabular}[c]{@{}l@{}}Single encoder  \end{tabular}
& \ch  & \xm  & 43.20 \\
\hline
\begin{tabular}[c]{@{}l@{}}Single encoder  \end{tabular}
& \xm  & \ch  & 38.36 \\
\hline
\begin{tabular}[c]{@{}l@{}}Single encoder  \\with  semantics\end{tabular}  
& \ch & \xm  & 39.23 \\
\hline
\begin{tabular}[c]{@{}l@{}}Multiple encoders  \end{tabular} 
& \ch  & \xm & 43.45 \\
\hline
\begin{tabular}[c]{@{}l@{}}Multiple encoders \\ with joint features \end{tabular}     
& \ch & \xm & 52.60  \\ 
\bottomrule
\end{tabular}
\end{table}
For evaluating the performance, we use the intersection-over-union~\cite{everingham2010pascal} metric as is commonly used by similar methods such as \cite{chen2021moving}. The evaluation code is taken directly from SemanticKITTI. We evaluate our model in the BEV space since most of the algorithms downstream from the perception task, such as path planning, are carried out in the BEV space. We report Intersection over Union (IoU) for moving class.

To prove the real-time capabilities of our proposed architecture, we run the inference on an Nvidia Jetson Xavier AGX development kit, a commonly used automotive embedded systems platform for deep learning applications. As is shown in Table \ref{tab:kpibenchmark}, we achieve an impressive inference latency of 8ms (inference speed of 125 Frames Per Second (FPS)). In terms of accuracy, we are slightly behind SalsaNext \cite{cortinhal2020salsanext}. We haven't performed extensive hyperparameter tuning or data augmentation.\par
\begin{table}[t]
\centering
\caption{\bf Ablation study of inference models illustrating the precision, accuracy, and size.}
\label{tab:ablation-reducedprec}
\begin{tabular}{@{}llcl@{}}
\toprule
\textit{Precision} & \textit{IoU} & \textit{Latency (ms)} & \textit{Size (MB)} \\ \midrule
FP32               & 52.60         & 8                     & 35.0                 \\
FP16               & 51.40         & 3                     & 15.3               \\
INT8               & 48.07        & 2                     & 8.0                  \\ \bottomrule
\end{tabular}
\end{table}
\subsection{Ablation Study}

We perform extensive ablation studies using different modifications to our architecture and input data representation.
As can be seen in Table \ref{tab:ablation1}, we evaluate using both a single encoder-decoder network as well as with multiple encoders and joint decoder architecture, which is more common in optical flow-based approaches such as \cite{ilg2017flownet}. However, surprisingly enough, the single encoder-decoder architecture achieves better accuracy than a multi-encoder approach where joint features were not computed. We explain that each encoder during training learns features local to its input BEV frame for the multi-encoder approach. However, due to the absence of any DB blocks for joint feature learning, sufficient joint features are not learned, which leads to a lower representation of motion in feature space.

We also experiment with adding full semantic segmentation masks to individual input BEVs. This, however, does not seem to offer any benefits. The network seems to be biased towards classifying all instances of objects as positive for motion in this case. This could be directly attributed to the full semantic masks that do not differentiate between mobile and static objects.
Furthermore, we evaluate another commonly used residual computation approach like subtraction and find that the proposed approach to multiply performs best.


We experiment with reduced precision inference as shown in Table \ref{tab:ablation-reducedprec} at 16-bit floating point (FP16) and 8-bit integer (INT8) and demonstrate that performance is not degraded by a large amount even at a reduced precision. This is directly relevant for real-time embedded system based deployment.
\section{CONCLUSION}

In this paper, we demonstrated an algorithm that incrementally builds upon an existing network for object detection to do motion segmentation so that it can be added as an additional task in a multi-task learning framework. We proposed a residual computation layer that exploits the disparity between the static and mobile parts of two successive motion-compensated frames. We also proposed a data augmentation technique that greatly improves the class imbalance between static and mobile points present in the SemanticKITTI dataset. We observed that motion segmentation in BEV space is not a straightforward task due to the sparsity of the 3D points and a relatively small cross-section of several traffic elements like pedestrians and bicyclists. \\

\noindent \textbf{ACKNOWLEDGEMENT}
\vspace{2mm}
\begin{wrapfigure}{l}{0.01\textwidth}
\includegraphics[width=7mm]{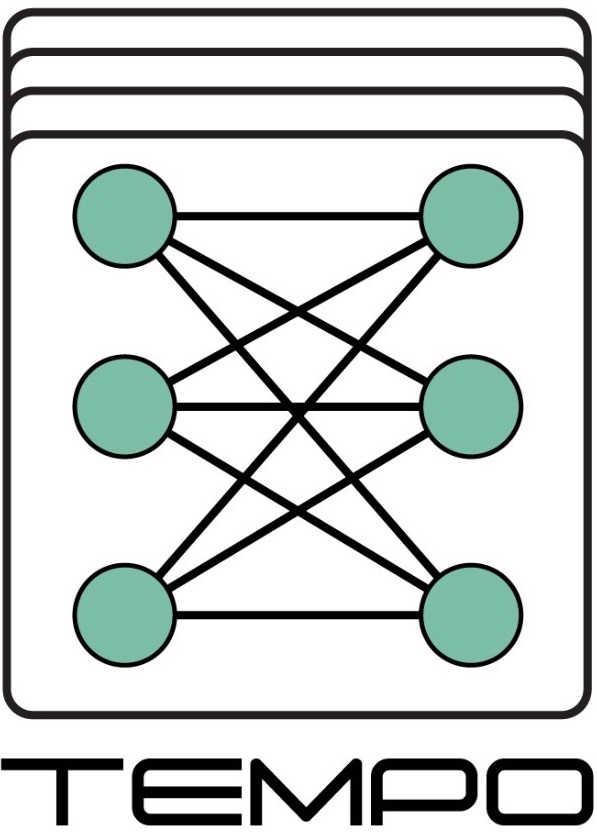}
\end{wrapfigure}
We are funded by the Electronic Components and Systems for European Leadership Joint Undertaking grant No 826655 receiving support from the European Union’s Horizon 2020 research and innovation programme. Further partial funding is provided by the German Federal Ministry of Education and Research.

\bibliographystyle{apalike}
{\small
\bibliography{references.bib}}
\end{document}